\documentclass{article}

\usepackage{microtype}
\usepackage{graphicx}
\usepackage{booktabs} % for professional tables

\usepackage{hyperref}
\usepackage{amsfonts}

\usepackage[accepted]{icml2021}
\usepackage[labelsep=quad,indention=10pt]{subfig}
\usepackage{amsmath}
\DeclareMathOperator*{\argmax}{argmax}

\icmltitlerunning{Learning who is in the market from time series}

\begin{document}

\twocolumn[
\icmltitle{Learning who is in the market from time series: market participant discovery through adversarial calibration of multi-agent simulators.}



\icmlsetsymbol{equal}{*}

\begin{icmlauthorlist}
\icmlauthor{Victor Storchan}{too}
\icmlauthor{Svitlana Vyetrenko}{to}
\icmlauthor{Tucker Balch}{to}
\end{icmlauthorlist}

\icmlaffiliation{to}{JP Morgan AI Research, New York, USA}
\icmlaffiliation{too}{JP Morgan, Palo Alto, USA}

\vskip 0.1in

\icmlcorrespondingauthor{Victor Storchan}{victor.storchan@jpmchase.com}
\icmlcorrespondingauthor{Svitlana Vyetrenko}{svitlana.s.vyetrenko@jpmchase.com}
\icmlcorrespondingauthor{Tucker Balch}{tucker.balch@jpmchase.com}
\icmlkeywords{Multi-agent systems, generative adversarial networks, time series}

\vskip 0.1in
]
\printAffiliationsAndNotice{}



%UNCOMMENT!!!

\vspace{-1cm}
\begin{abstract}
\small{
In electronic trading markets often only the price or volume time series, that result from interaction of multiple market participants, are directly observable. In order to test trading strategies before deploying them to real-time trading, multi-agent market environments calibrated so that the time series that result from interaction of simulated agents resemble historical are often used. To ensure adequate testing, one must test trading strategies in a variety of market scenarios -- which includes both scenarios that represent ordinary market days as well as stressed markets (most recently observed due to the beginning of COVID pandemic). In this paper, we address the problem of multi-agent simulator parameter calibration to allow simulator capture characteristics of different market regimes.  We propose a novel two-step method to train a discriminator that is able to distinguish between “real”  and  “fake” price and volume time series as  a  part of GAN with self-attention, and then utilize it within an optimization framework to tune parameters of a simulator model with known agent archetypes to represent a market scenario. We conclude with experimental results that demonstrate effectiveness of our method.}
\end{abstract}

\vspace{-1cm}

\section{Introduction}
Multi-agent simulation is widely used in financial markets to study macroscopic structural effects in counterfactual  scenarios. It is especially useful as a natural bottom-up approach in situations where the environment needs to be responsive to experimental agent's actions such as trading strategy testing. Thorough trading strategy testing must include exposure to a variety of market scenarios in order to gain understanding how experimental agents will behave when market dynamics shifts. For example, in March 2020 markets became significantly more volatile and illiquid due to the start of COVID pandemic, which affected performance of trading strategies. Therefore, there is a practical need to be able to calibrate multi-agent simulators to a variety of scenarios for proper risk management.

Explainability is an additional benefit of multi-agent simulation. One can provide insightful answers to "what if" questions by explicitly changing agent composition of simulated markets. When individual agent-specific data is available to the researcher, it can be used for the learning the market agent policy (e.g., \cite{VyetrenkoDecisionTrees}, \cite{nasdaq}); however, most frequently such labeled data is proprietary and only the price or volume time series that result from interaction of multiple market participants are directly observable. Some market participants, such as exchanges or over-the-counter market markers, however, have additional knowledge of parametrized agent archetypes that play in the market (momentum traders, value traders, noise (retail) traders, etc.) -- such participants are interested in tuning parameters of the known agent archetypes so that the result of their interaction resembles historical time series for counterfactual scenario calibration. Explicit optimization objectives for simulator parameter tuning are, however, difficult to construct due the complex nature of time series that results from agent interaction. Therefore,  we propose  to  learn  a  discriminator  (i.e.   a  classifier)  that  can distinguish synthetic time series from real,  and use it as an implicitly written objective function for the calibration task.

\vspace{-0.2cm}
\section{Our contributions}
\begin{figure}[t!]
\centering
\includegraphics[scale=0.4]{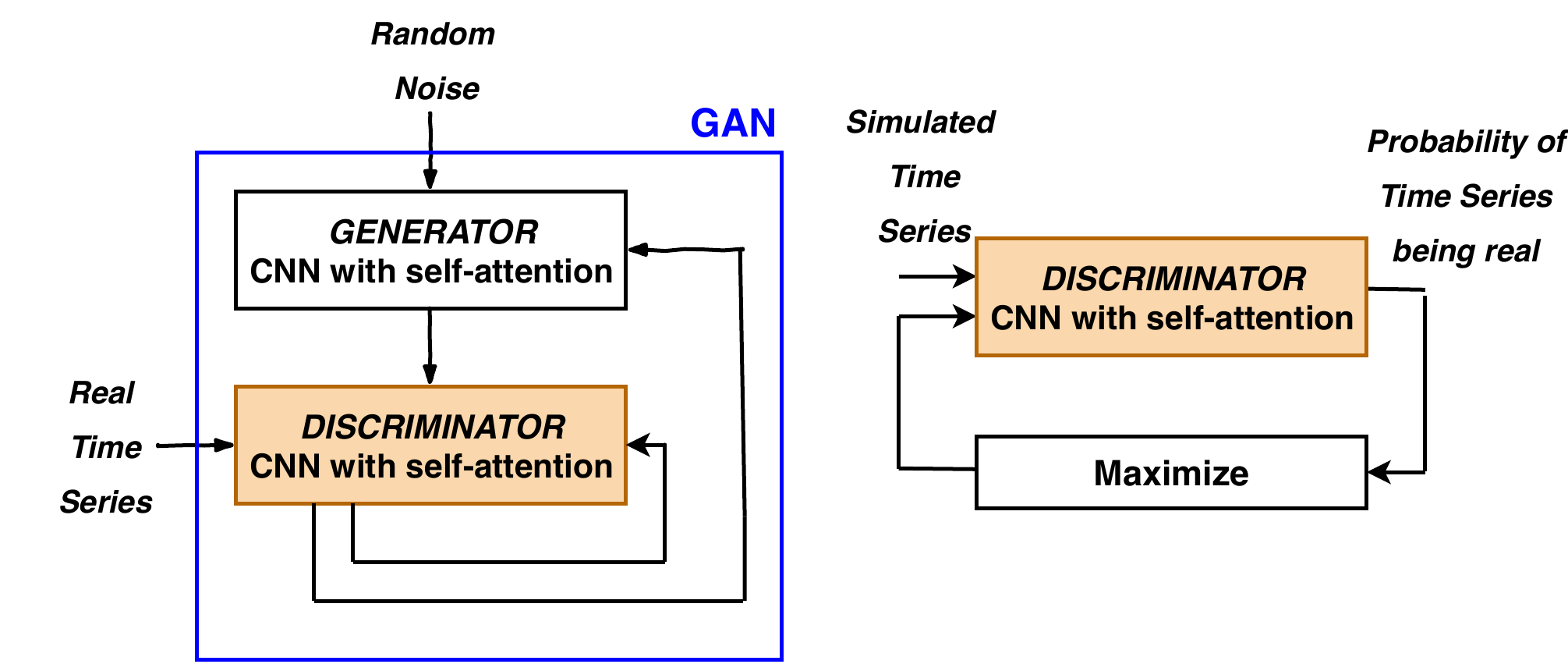}
\caption{MAS-GAN approach schematic.}
\label{fig:gan_chart}
\end{figure}

In this paper, we propose MAS-GAN -- a two-step method for multi-agent market simulator calibration. A discriminator is trained in competition with the generator as a part of GAN to distinguish synthetic time series from real (i.e. historical). Both generator and discriminator are enhanced by self-attention layers for better time series resolution quality. 
%At training time, we use the Wasserstein GAN framework which is more stable \cite{arjovsky2017wasserstein}. 
The input of trained discriminator is a synthetic time series, and the output uses sigmoid activation which allows to interpret it as probability of synthetic time series being real. Similar to \cite{liang2018enhancing}, the discriminator score is higher if the synthetic time series shares more features with the historical dataset. One can then use the discriminator score as an implicit optimization objective, and optimize simulated model parameters to determine the set that produces most realistic time series (see Figure~\ref{fig:gan_chart} for schematic).  

To the best of our knowledge, MAS-GAN is the first method that proposes to use adversarially trained discriminator as an objective function for multi-agent system optimization. Unlike \cite{modelAssistedGAN}, our method is model-agnostic and does not require learning a simulator approximation which can be expensive. Because MAS-GAN learns from historical dataset and is unbiased toward synthetic data coming from the simulator, once the discriminator is trained -- it can be used to calibrate any time series model (not necessarily multi-agent) that is intended to model that historical dataset. %Since simulated data is not used for GAN training, the discriminator is not biased to any particular model and only uses historical data as a 'ground truth' for learning -- this provides an additional justification of a two-step as opposed to one-step calibration method. 
In contrast with \cite{li2020stockgan}, our method retains full explainability of multi-agent simulations as GANs in our case are not used for black-box data generation, but rather only for learning a calibration objective given by the discriminator.

Let $S_{R}(v)$ be the simulated mid price time series obtained using the parameter vector $v$ and random seed $R$.  Since the discriminator $D$ outputs probability of time series $S_R(v)$ being real, we are interested in finding parameter vector $v^*$ that maximizes this probability. Formally, we are interested in finding $v^* = \argmax_v \mathbb{E}_R \left[ D( S_{R}(v) ) \right]$. We then perform grid-based optimization with respect to the discriminator-defined objective to find most realistic simulation parameters out of the given grid. %We note here that our approach does not attempt to improve the simulated model but rather finds a set of parameters such that the existing model describes $P_{real}$ most closely.
%\begin{eqnarray}
%v^* = \argmax_{v} \mathbb{E}_R \left[ D( S_{R}(v) ) \right].
%\label{optim}
%\end{eqnarray} 

\vspace{-0.2cm}
\section{Training details}
Most of the GAN literature focuses on training GANs for the purpose of subsequently using only data generator, and discriminator is only viewed as an auxiliary agent that assists with generator training. To complement this, using GAN-trained discriminator for the purpose of time series anomaly detection is discussed in \cite{schlegl2017unsupervised}, \cite{mattia2019survey}, \cite{GAN_anomaly}.%, \cite{tomatoGAN}. 
By learning statistical similarities between non-anomalous time series, one can learn to determine what constitutes an anomaly. In this paper, we rely on a similar idea and introduce an application of adversarially trained discriminators to multi-agent simulator parameter optimization.

Using GANs to generate synthetic time series has been widely studied -- however, only LSTMs and recurrent neural network architectures have been previously considered (e.g., \cite{timeseriesGAN}). To the best of our knowledge, we are innovating by using both conv1D and self-attention without recurrence in the architecture to generate synthetic time series. We use conv1D to encode the local correlations in the data. The deeper the network is, the more global encoding will happen. We add self-attention to provide a more targeted way to encode global correlations with a single layer. A 2D version of the self-attention layer was described in \cite{Sagan}. For this work, we use a similar implementation of a 1D version of the self-attention layer (see Figure~\ref{fig:self_attention} in Appendix~\ref{self_attention_appendix} for schematic).
\vspace{-0.2cm}
\subsection{Self-Attention}
Transformer's head mechanisms, introduced in \cite{transformers} have proved their efficiency at capturing global dependencies in multiple domains of applications such as translation, language modeling or object detection. Among the different attention mechanisms, self attention is computing a representation of the next position in a sequence by looking at all the position in the sequence and learning regions of interest. In self-attention, we route information according to the information itself, so according to the incoming information. We do that by expressing queries expressing the kind of information needed from the previous layer and keys exposing the kind of information the nodes of the next layer are containing. To aggregate the results, we use the softmax function which is a kind of soft routing: all the nodes of the previous layer are contributing to the expression of the node of the next layer but the majority goes to the places where the inner product is high (see Figure~\ref{fig:self_attention} in Appendix~\ref{self_attention_appendix}).
\vspace{-0.2cm}
\subsection{WGAN with gradient penalty}
In the original paper \cite{goodfellow2014generative}, GANs are trained using the Jensen-Shannon Divergence (JSD), essentially to solve the problem of KL divergence in Variationnal Autoencoders (VAE) which is note define in the real world scenario where the model manifold and the true distribution’s support have not a non-negligible intersection \cite{arjovsky2017principled}. However, in \cite{arjovsky2017wasserstein} example 1 shows that it is still not a good notion of distance. The authors introduce Wasserstein GANs which use the earth mover distance as a distance between distribution. It has the advantage of being a true metric: a measure of distance in a space of probability distributions. Instead of having a discriminator evaluating the probability of a sample to be real, we train a critics that is computing a realism score for each sample. WGANs are less vulnerable to getting stuck than minimax-based GANs, and avoid problems with vanishing gradients. However, the Kantorovitch-Rubinstein duality is stating that the critics should lie in the set of K-lipschitz functions. To implement this constraint, the original WGAN paper is clipping the weights of the critics. But it biases it towards learning much simpler functions. We chose to implement the constraint proposed by \cite{gulrajani2017improved} where the norm of gradient of the critic with respect to its input is penalized. See Figure~\ref{fig:gan_discr} in Appendix~\ref{appendix_GAN} for the detailed schematic of GAN architecture.

\vspace{-0.2cm}
\section{Experimental results}
\subsection{Simulated environment}
\label{sim_env}
\begin{figure}[t!]
 \centering
 \includegraphics[width=0.3\textwidth]{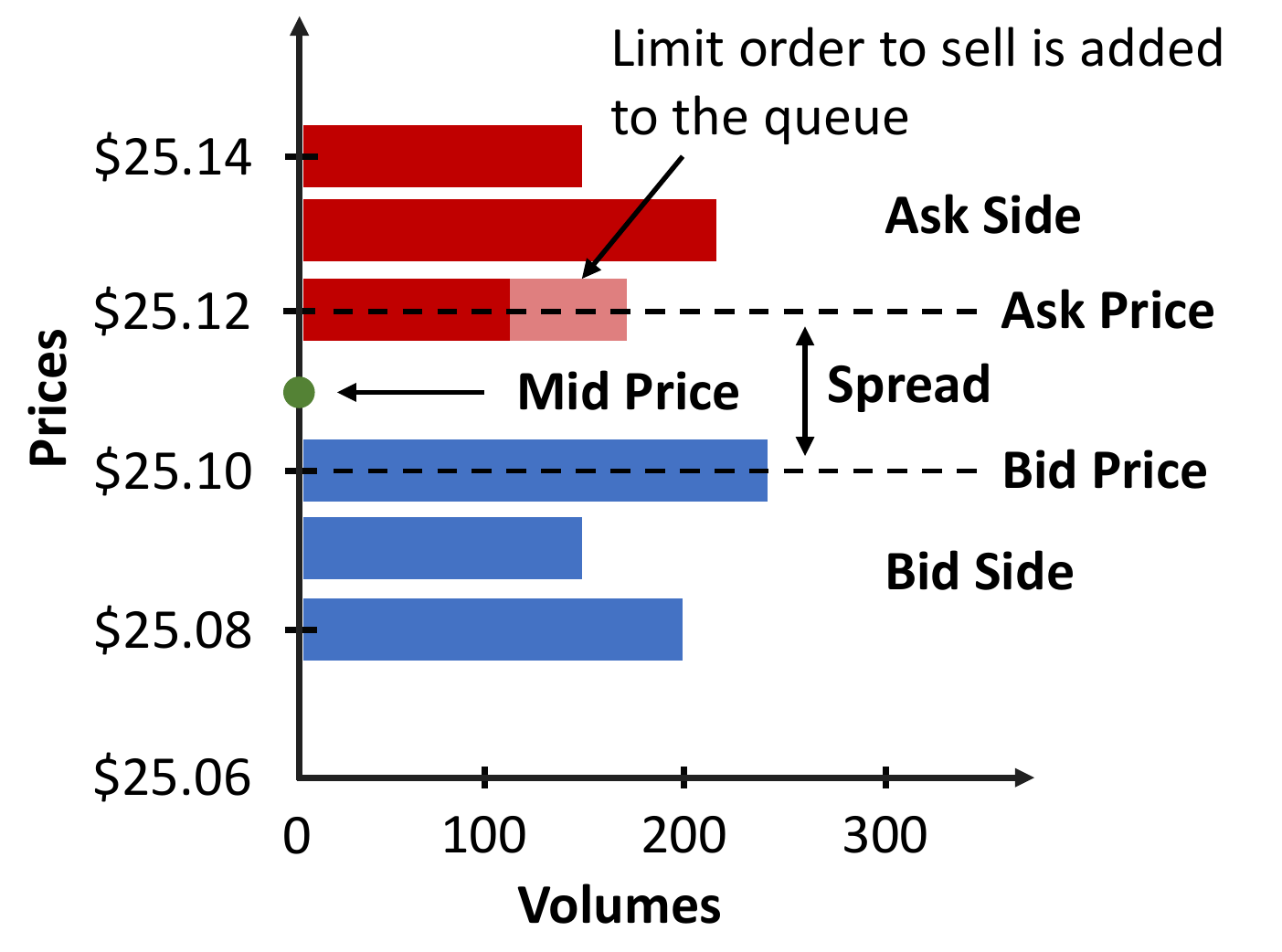}
 \caption{Visualization of the LOB structure.}
 \vspace{-1cm}
 \label{fig:LOB}
 \end{figure}

 Public exchanges and over-the-counter market makers facilitate the buying and selling of assets by accepting and satisfying buy and sell orders from multiple market participants via limit order book (LOB) mechanism, which is an electronic record of the queues of outstanding buy and sell limit orders organized by price levels \cite{Bouchaud_book}. See Figure~\ref{fig:LOB} for visualization of the LOB structure. 
 
 We apply MAS-GAN to calibrate the LOB simulation environment [citation redacted] that provides a selection of basic agent archetypes (such as market makers, noise agents, value agents, etc.), a NASDAQ-like exchange agent which lists any number of securities for trade against an LOB with price-then-FIFO matching rules, and a simulation kernel which manages the flow of time and handles all inter-agent communication. 
 
We use agent archetype implementation introduced in~\cite{vyetrenko2019get}. {\it{Value agents}} are designed to simulate
the actions of fundamental traders that trade according
to their belief of the exogenous value of a stock (also called fundamental price -- which in our model follows
 a mean reverting price time series). Value traders arrive to the market at a Poisson rate $\lambda$ and choose to buy or sell a stock
depending on whether it is cheap or expensive relative to their
noisy observation of a fundamental price. The fundamental follows a discrete-time mean-reverting Ornstein-Uhlenbeck process~\cite{byrd2019explaining}. {\it{Noise agents}} are designed to emulate the
actions of retail players who trade on demand without
any other considerations; $N$ noise agents arrive to the market at times that are uniformly distributed
throughout the trading day and place an order of random size in random direction. Finally, {\it{market makers}} act as liquidity providers by placing
limit orders on both sides of the LOB with a constant arrival rate. 
\vspace{-0.2cm}
\subsection{Training details}
In real limit order book markets, multiple correlated data streams are produced as a result of participant interaction (e.g., prices and quotes at multiple limit order book levels and traded volumes). MAS-GAN allows to capture temporal autocorrelations and cross-correlations of multiple limit order book time series distributions. Specifically, we train a generative adversarial network on $T$-second mid price returns and cumulative $T$-second traded volumes. To produce the input vector that is passed to the GAN discriminator, we concatenate $T$-second mid price returns and volumes. It is done in order for the neural network with self-attention to encode correlations between price and volume time series and to retranscribe them when generating time series from noise. %Mode collapse, when generator output shows little to no variability regardless of the input, is a common problem in GAN training. To avoid the mode collapse, we rescaled the volumes so that their order of magnitude is similar to that of the mid price returns. 

We note that in order to train a high quality discriminator, the generator needs to be capable to generate a diverse set of realistic time series. We train the GAN iteratively until the following three conditions are satisfied: (1) Generated time series visually show diversity -- see Figure~\ref{fig:generator diversity} in Appendix~\ref{Supplementary materials}  for visual inspection. We are aware of the issues raised by \cite{overfit} that visual inspection might not be sufficient to demonstrate diversity as time series can be simply memorized by GANs and leave this question for further work; (2) Asset return distributions and volume/volatility correlation of the generated time series converge to historical  -- see Figure~\ref{fig:returns} in Appendix~\ref{Supplementary materials} for the training progression; (3) The discriminator cannot distinguish generated time series from real -- see Figure~\ref{fig:discr_scores} in Appendix~\ref{Supplementary materials} for distribution of discriminator scores during training.

%\begin{enumerate}
%    \vspace{-0.4cm}
%    \item Generated time series visually show diversity -- see Figure~\ref{fig:generator diversity} in Appendix~\ref{Supplementary materials}  for visual inspection. We are aware of the issues raised by \cite{overfit} that visual inspection might not be sufficient to demonstrate diversity as time series can be simply memorized by GANs and leave this question for further work. 
%    \vspace{-0.2cm}
%    \item Asset return distributions and volume/volatility correlation of the generated time series converge to historical  -- see Figure~\ref{fig:returns} in Appendix~\ref{Supplementary materials} for the training progression.
%    \vspace{-0.4cm}
%    \item The discriminator cannot distinguish generated time series from real -- see Figure~\ref{fig:discr_scores} in Appendix~\ref{Supplementary materials} for distribution of discriminator scores during training.
%\end{enumerate}

\begin{figure*}[htb!]
\centering
\subfloat[$N^*=5000$, $\lambda^*=\mathrm{e}{-14}$. Parameter configuration used to simulate liquid markets.]
{\includegraphics[width=0.3\textwidth]{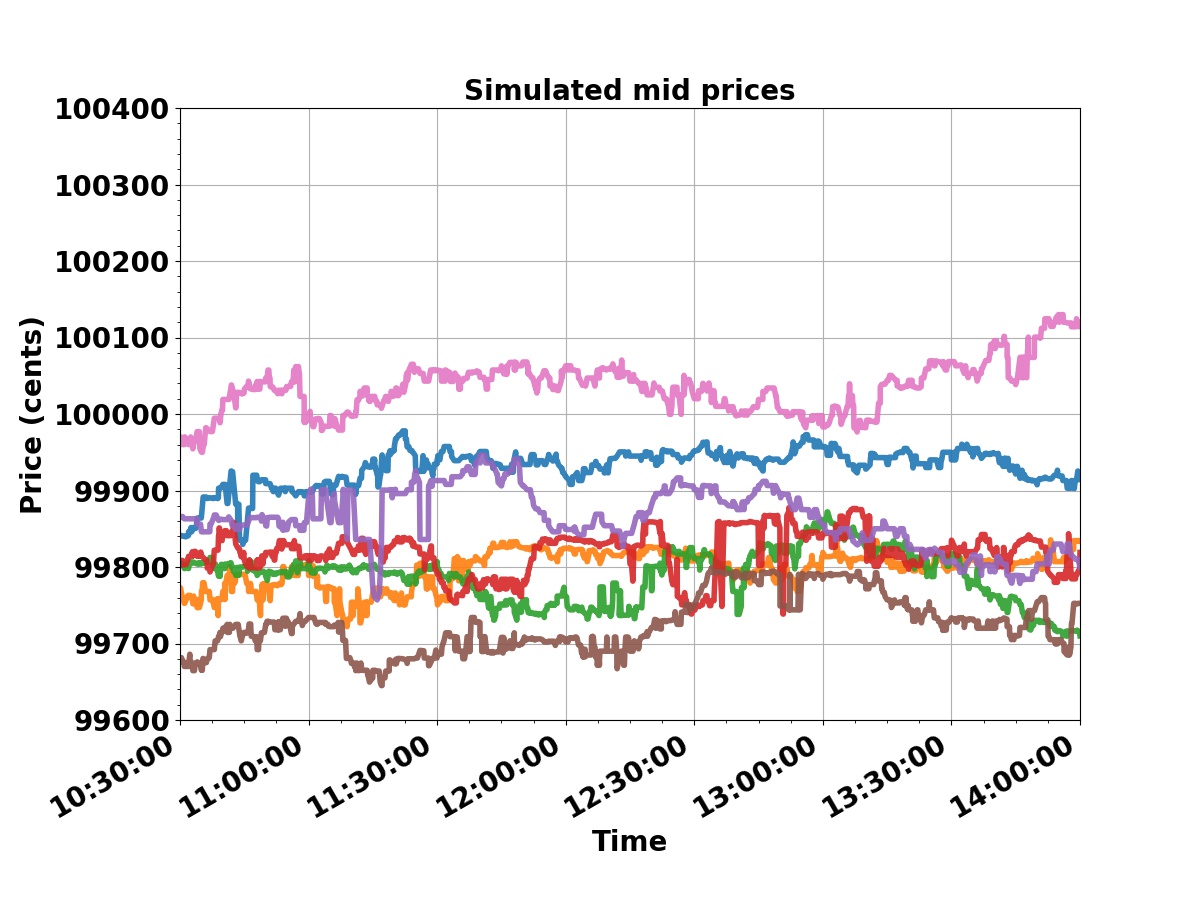}}
\subfloat[$N^*=4800$, $\lambda^*=\mathrm{e}{-13}$. Parameter configuration used to simulate illiquid markets.]
{\includegraphics[width=0.3\textwidth]{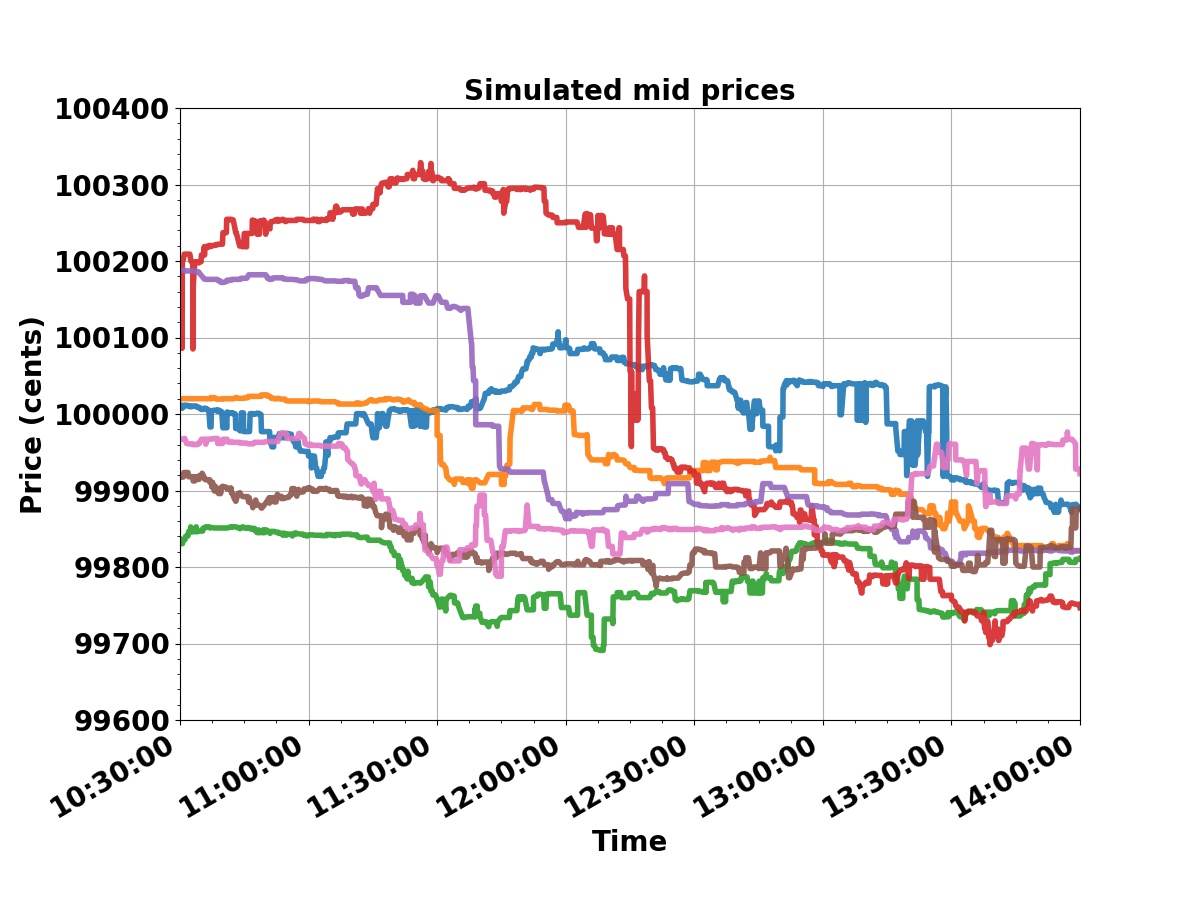}}
\caption{Mid price time series.}
\label{time_series_examples}
\end{figure*}

\begin{figure*}[htb!]
\centering
\subfloat[$N^*=5000$, $\lambda^*=\mathrm{e}{-14}$]
{\includegraphics[width=0.32\textwidth]{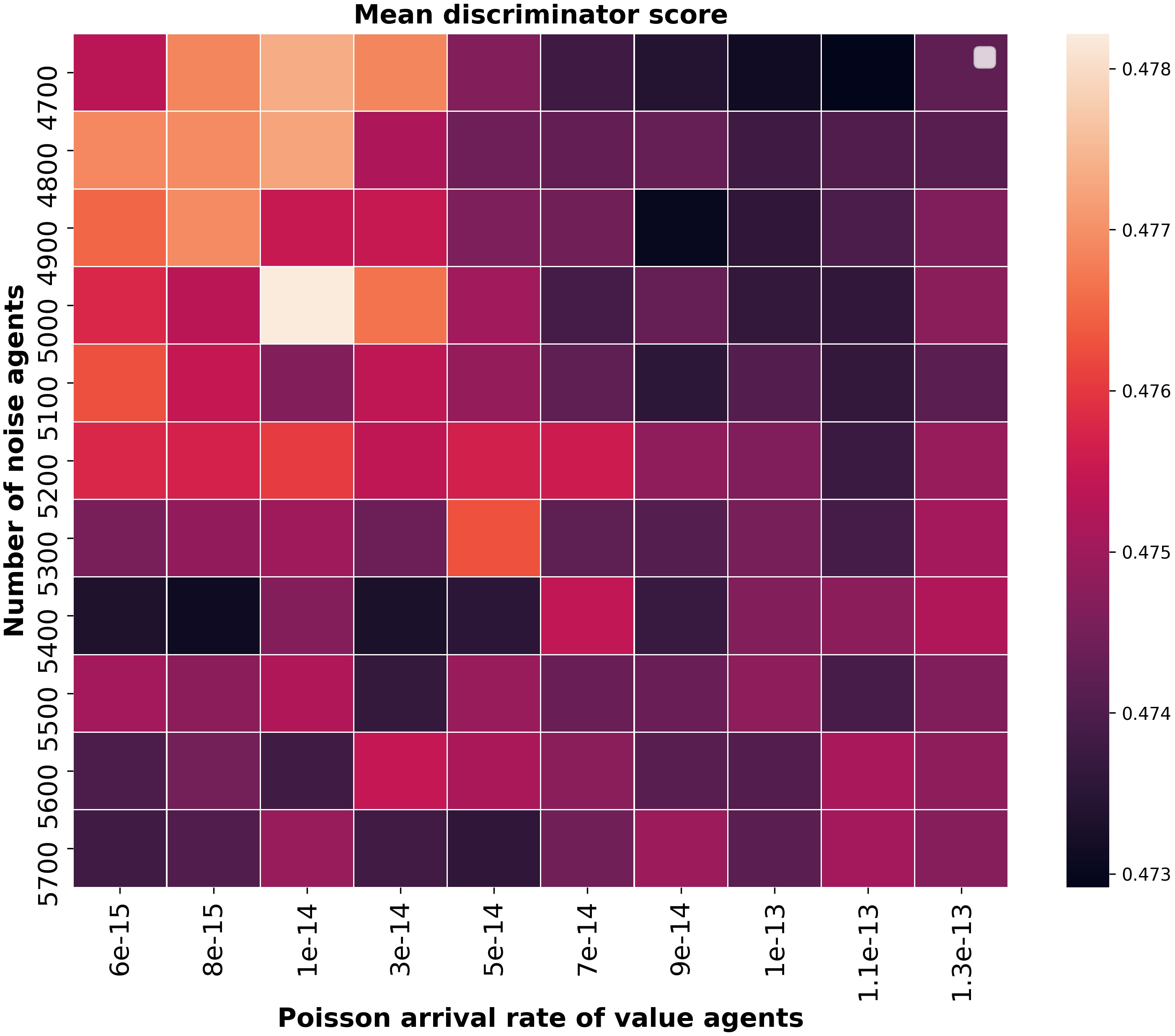}}
\subfloat[$N^*=4800$, $\lambda^*=\mathrm{e}{-13}$]
{\includegraphics[width=0.32\textwidth]{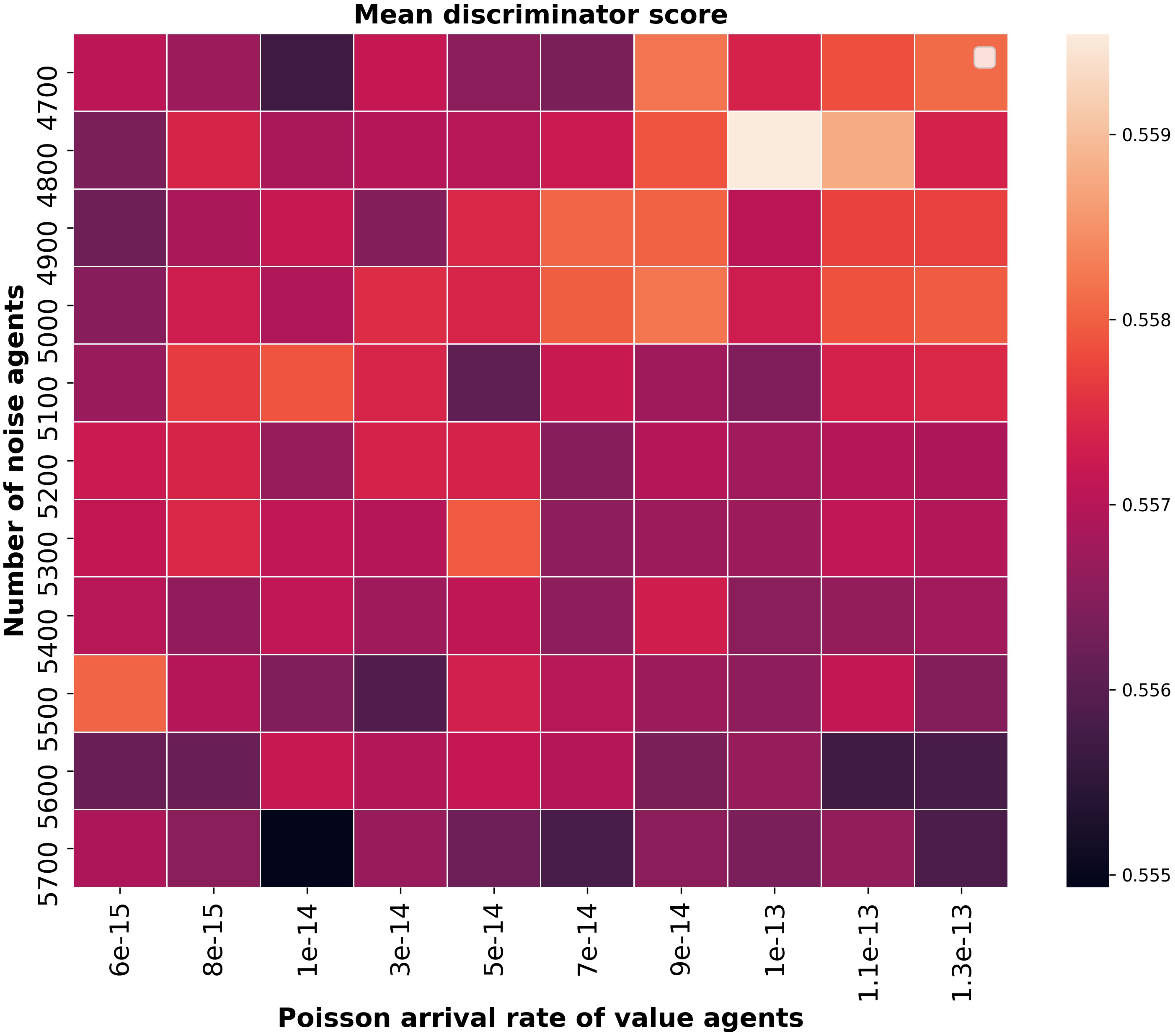}}
\caption{Mean discriminator score heatmap with respect to the noise agent number $N^*$ and value agent arrival rate $\lambda^*$. For each grid configuration, we run 20 simulations with different seeds for initialization of pseudo-random number generator.}
\label{fig:heatmap_reconstruct}
\end{figure*}

\subsection{Configuration 'recovery' experiment}
\label{numerical_experiment}
We test MAS-GAN on synthetic configurations that are designated as "ground truth". We consider a family of synthetic market configurations with a single market maker, $100$ value agents that arrive to the market at Poisson rate $\lambda$ and $N$ noise agents (see agent descriptions in Section~\ref{sim_env}) - with $\lambda$ and $N$ being calibration parameters. In Figure~\ref{time_series_examples}  one can see simulated mid price time series generated for different values of parameters $\lambda$ and $N$. Different parameter values can be used to model different liquidity regimes. For example, parameter values from Figure~\ref{time_series_examples}(a) can used to simulate "ordinary" liquid markets, whereas, parameter values from Figure~\ref{time_series_examples}(b) model a "bumpy" illiquid markets -- such markets can be can be observed during shock periods such as those that were seen in the beginning of COVID pandemic. We fix $\lambda^*$ and $N^*$ that correspond to different liquidity configurations as per charts Figure~\ref{time_series_examples}(a) and (b). For different seeds for initialization of pseudo-random number generator, one can produce multiple time series samples — that is, simulate trading days with fixed agent participation.  We run the following experiments to verify the validity of MAS-GAN: first, train a discriminator as a part of GAN with self-attention to recognize time series samples for fixed $\lambda^*$ and $N^*$ and varying random seeds; second,for each choice of $\lambda$ and $N$ over the grid, apply the trained discriminator to “recognize”  $\lambda^*$ and $N^*$ — a configuration that a discriminator was trained against. Figure~\ref{fig:heatmap_reconstruct} illustrates the outcome of the above  experiment for the respective values of $\lambda^*$ and $N^*$ as in Figure~\ref{time_series_examples}. We observe that applying discriminator to the rectangular grid can identify the configuration that the discriminator was trained against. We also notice that configurations in the immediate neighborhood of $\lambda^*$ and $N^*$ on the grid are identified as more likely to have trained the discriminator than those that are further away.

\vspace{-0.3cm}
\section{Conclusion}
In summary, we propose a novel GAN-based method for multi-agent simulator parameter calibration and demonstrate its effectiveness experimentally. Our method relies on the basic knowledge of market agent archetypes, and can be effectively used for agent micro-parameter tuning to different market regimes by market participants such as exchanges or market makers to whom knowledge of basic archetypes is available.

\vspace{-0.3cm}
\section{Acknowledgements} 
\small{
The authors would like to thank Alexia Jolicoeur Martineau for the insightful discussion on the relevance of using a trained discriminator to evaluate realism of samples. This paper was prepared for informational purposes [“in part” if the work is collaborative with external partners] by the Artificial Intelligence Research group of JPMorgan Chase and Co and its affiliates (“J.P. Morgan”), and is not a product of the Research Department of J.P. Morgan.  J.P. Morgan makes no representation and warranty whatsoever and disclaims all liability, for the completeness, accuracy or reliability of the information contained herein.  This document is not intended as investment research or investment advice, or a recommendation, offer or solicitation for the purchase or sale of any security, financial instrument, financial product or service, or to be used in any way for evaluating the merits of participating in any transaction, and shall not constitute a solicitation under any jurisdiction or to any person, if such solicitation under such jurisdiction or to such person would be unlawful.}
\bibliography{icml_masgan}

\begin{thebibliography}{20}
\providecommand{\natexlab}[1]{#1}
\providecommand{\url}[1]{\texttt{#1}}
\expandafter\ifx\csname urlstyle\endcsname\relax
  \providecommand{\doi}[1]{doi: #1}\else
  \providecommand{\doi}{doi: \begingroup \urlstyle{rm}\Url}\fi

\bibitem[Alonso{-}Monsalve \& Whitehead(2018)Alonso{-}Monsalve and
  Whitehead]{modelAssistedGAN}
Alonso{-}Monsalve, S. and Whitehead, L.~H.
\newblock Image-based model parameter optimisation using model-assisted
  generative adversarial networks.
\newblock \emph{CoRR}, 2018.
\newblock URL \url{http://arxiv.org/abs/1812.00879}.

\bibitem[Arjovsky \& Bottou(2017)Arjovsky and Bottou]{arjovsky2017principled}
Arjovsky, M. and Bottou, L.
\newblock Towards principled methods for training generative adversarial
  networks, 2017.

\bibitem[Arjovsky et~al.(2017)Arjovsky, Chintala, and
  Bottou]{arjovsky2017wasserstein}
Arjovsky, M., Chintala, S., and Bottou, L.
\newblock Wasserstein gan, 2017.

\bibitem[Bouchaud et~al.(2018)Bouchaud, Bonart, Donier, and
  Gould]{Bouchaud_book}
Bouchaud, J.-P., Bonart, J., Donier, J., and Gould, M.
\newblock \emph{{Trades, quotes and prices: financial markets under the
  microscope}}.
\newblock Cambridge University Press, Cambridge, 2018.

\bibitem[Byrd(2019)]{byrd2019explaining}
Byrd, D.
\newblock Explaining agent-based financial market simulation.
\newblock \emph{arXiv preprint arXiv:1909.11650}, 2019.

\bibitem[Darley \& Outkin(2007)Darley and Outkin]{nasdaq}
Darley, V. and Outkin, A.~V.
\newblock \emph{{A NASDAQ Market Simulation:Insights on a Major Market from the
  Science of Complex Adaptive Systems}}.
\newblock World Scientific Publishing Co. Pte. Ltd., December 2007.

\bibitem[Esteban et~al.(2017)Esteban, Hyland, and Rätsch]{timeseriesGAN}
Esteban, C., Hyland, S., and Rätsch, G.
\newblock Real-valued (medical) time series generation with recurrent
  conditional gans.
\newblock 06 2017.

\bibitem[Goodfellow et~al.(2014)Goodfellow, Pouget-Abadie, Mirza, Xu,
  Warde-Farley, Ozair, Courville, and Bengio]{goodfellow2014generative}
Goodfellow, I.~J., Pouget-Abadie, J., Mirza, M., Xu, B., Warde-Farley, D.,
  Ozair, S., Courville, A., and Bengio, Y.
\newblock Generative adversarial networks, 2014.

\bibitem[Gulrajani et~al.(2017)Gulrajani, Ahmed, Arjovsky, Dumoulin, and
  Courville]{gulrajani2017improved}
Gulrajani, I., Ahmed, F., Arjovsky, M., Dumoulin, V., and Courville, A.
\newblock Improved training of wasserstein gans, 2017.

\bibitem[Li et~al.(2018)Li, Chen, Goh, and Ng]{GAN_anomaly}
Li, D., Chen, D., Goh, J., and Ng, S.
\newblock Anomaly detection with generative adversarial networks for
  multivariate time series.
\newblock \emph{CoRR}, abs/1809.04758, 2018.
\newblock URL \url{http://arxiv.org/abs/1809.04758}.

\bibitem[Li et~al.(2020)Li, Wang, Lin, Sinha, and Wellman]{li2020stockgan}
Li, J., Wang, X., Lin, Y., Sinha, A., and Wellman, M.~P.
\newblock Generating realistic stock market order streams.
\newblock In \emph{To Appear in the 34th AAAI Conference on Artificial
  Intelligence}, 2020.

\bibitem[Liang et~al.(2018)Liang, Li, and Srikant]{liang2018enhancing}
Liang, S., Li, Y., and Srikant, R.
\newblock Enhancing the reliability of out-of-distribution image detection in
  neural networks.
\newblock In \emph{International Conference on Learning Representations}, 2018.
\newblock URL \url{https://openreview.net/forum?id=H1VGkIxRZ}.

\bibitem[Mattia et~al.(2019)Mattia, Galeone, Simoni, and
  Ghelfi]{mattia2019survey}
Mattia, F.~D., Galeone, P., Simoni, M.~D., and Ghelfi, E.
\newblock A survey on gans for anomaly detection, 2019.

\bibitem[Ryan et~al.(2019)Ryan, Rabin, Simon, and Jurie]{overfit}
Ryan, W., Rabin, J., Simon, L., and Jurie, F.
\newblock Detecting overfitting of deep generative networks via latent
  recovery.
\newblock \emph{Proceedings of the IEEE Conference on Computer Vision and
  Pattern Recognition}, 2019.
\newblock URL
  \url{https://openaccess.thecvf.com/content_CVPR_2019/papers/Webster_Detecting_Overfitting_of_Deep_Generative_Networks_via_Latent_Recovery_CVPR_2019_paper.pdf}.

\bibitem[Schlegl et~al.(2017)Schlegl, Seeböck, Waldstein, Schmidt-Erfurth, and
  Langs]{schlegl2017unsupervised}
Schlegl, T., Seeböck, P., Waldstein, S.~M., Schmidt-Erfurth, U., and Langs, G.
\newblock Unsupervised anomaly detection with generative adversarial networks
  to guide marker discovery, 2017.

\bibitem[Vaswani et~al.(2017)Vaswani, Shazeer, Parmar, Uszkoreit, Jones, Gomez,
  Kaiser, and Polosukhin]{transformers}
Vaswani, A., Shazeer, N., Parmar, N., Uszkoreit, J., Jones, L., Gomez, A.~N.,
  Kaiser, L., and Polosukhin, I.
\newblock Attention is all you need.
\newblock \emph{Advances in neural information processing systems}, 2017.
\newblock URL \url{https://arxiv.org/abs/1706.03762}.

\bibitem[Vyetrenko \& Xu(2019)Vyetrenko and Xu]{VyetrenkoDecisionTrees}
Vyetrenko, S. and Xu, S.
\newblock Risk-sensitive compact decision trees for autonomous execution in
  presence of simulated market response.
\newblock In \emph{ICML 2019 Workshop on AI in Finance}, 06 2019.

\bibitem[Vyetrenko et~al.(2020)Vyetrenko, Byrd, Petosa, Mahfouz, Dervovic,
  Veloso, and Balch]{vyetrenko2019get}
Vyetrenko, S., Byrd, D., Petosa, N., Mahfouz, M., Dervovic, D., Veloso, M., and
  Balch, T.~H.
\newblock Get real: Realism metrics for robust limit order book market
  simulations.
\newblock In \emph{International Conference on Artificial Intelligence in
  Finance}, 2020.

\bibitem[Wang et~al.(2016)Wang, Yan, and Oates]{TimeSeriesClass}
Wang, Z., Yan, W., and Oates, T.
\newblock Time series classification from scratch with deep neural networks:
  {A} strong baseline.
\newblock \emph{CoRR}, 2016.
\newblock URL \url{http://arxiv.org/abs/1611.06455}.

\bibitem[Zhang et~al.(2019)Zhang, I., Metaxas, and Odena]{Sagan}
Zhang, H., I., G., Metaxas, D., and Odena, A.
\newblock Self-attention generative adversarial networks.
\newblock \emph{International Conference on Machine Learning}, 2019.
\newblock URL \url{https://arxiv.org/pdf/1805.08318.pdf}.

\end{thebibliography}
\bibliographystyle{icml2021}

\appendix
\section{Appendix} 
\label{Supplementary materials}
\subsection{Evaluating the quality of generated time series}
\begin{figure*}[t!]
\centering
%\vspace*{-20cm}
\includegraphics[scale=0.4]{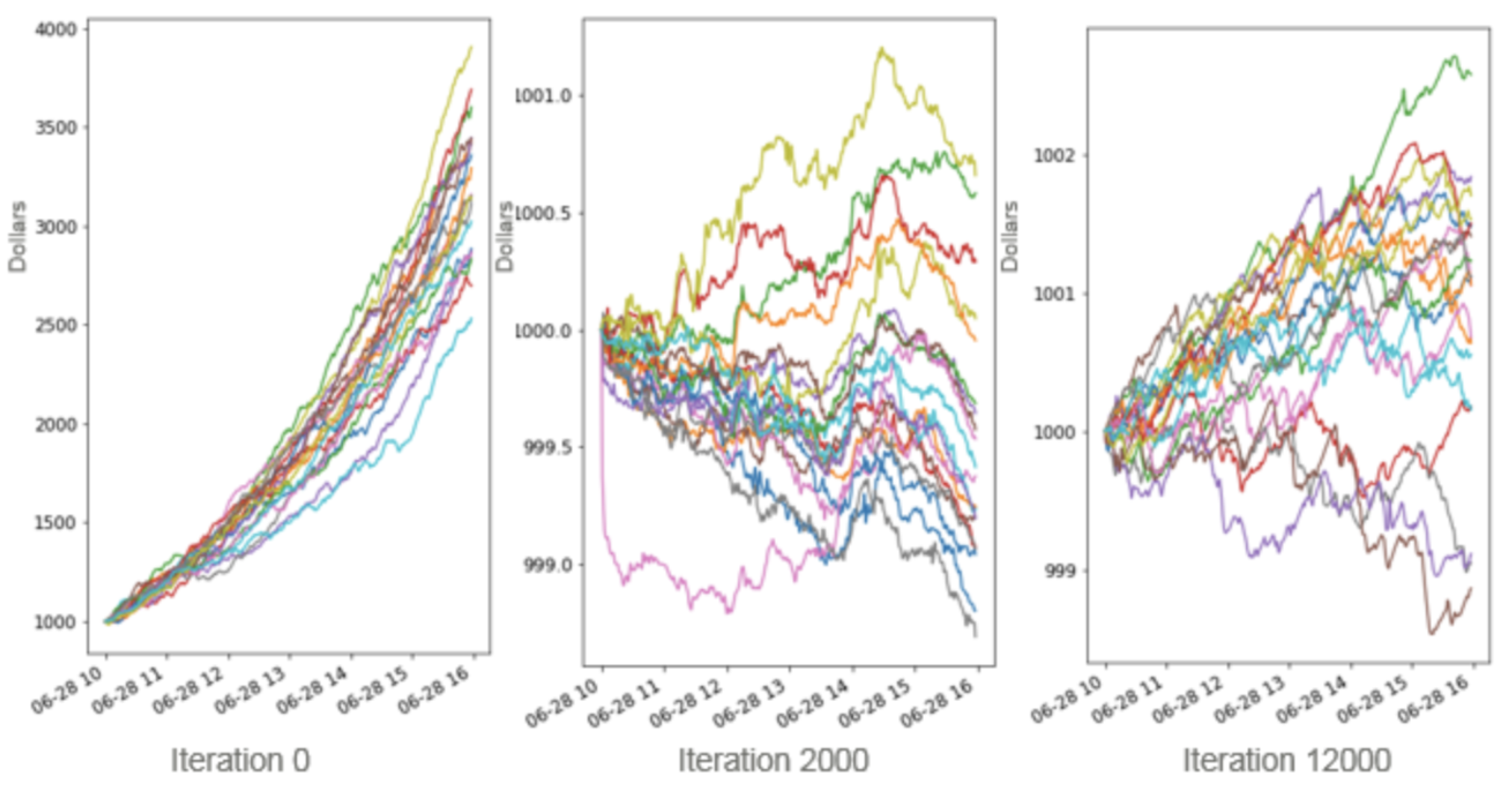}
\caption{Training progression of generated mid price returns. We visually observe diversity of generated price time series with training progression.}
\label{fig:generator diversity}
\end{figure*}
\begin{figure*}[t!]
\centering
\hspace*{-1cm}  
\subfloat[Initial.]
{\includegraphics[width=0.35\textwidth]{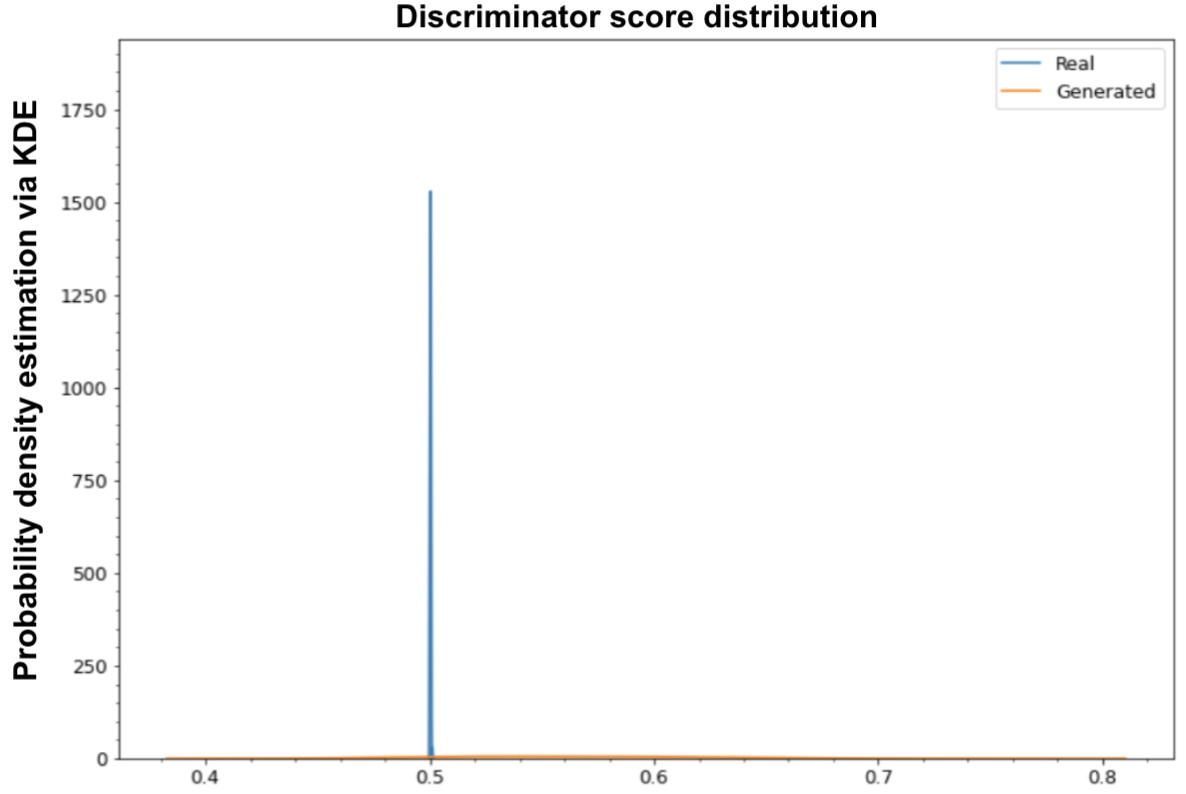}\label{fig:discr_2}}
\subfloat[After 5000 iterations.]
{\includegraphics[width=0.35\textwidth]{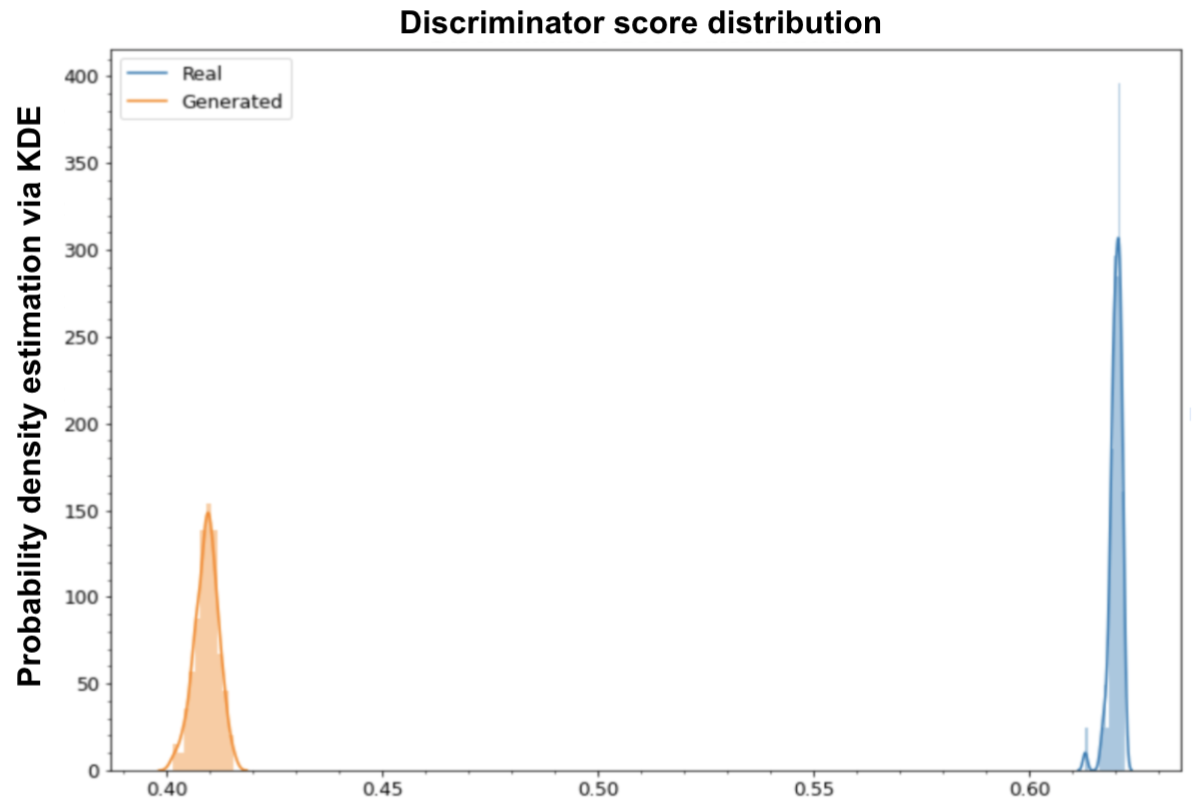}\label{fig:discr_3}}
\subfloat[After 10000 iterations. Note that 'real' and 'generated' distributions are overlapping.]
{\includegraphics[width=0.35\textwidth]{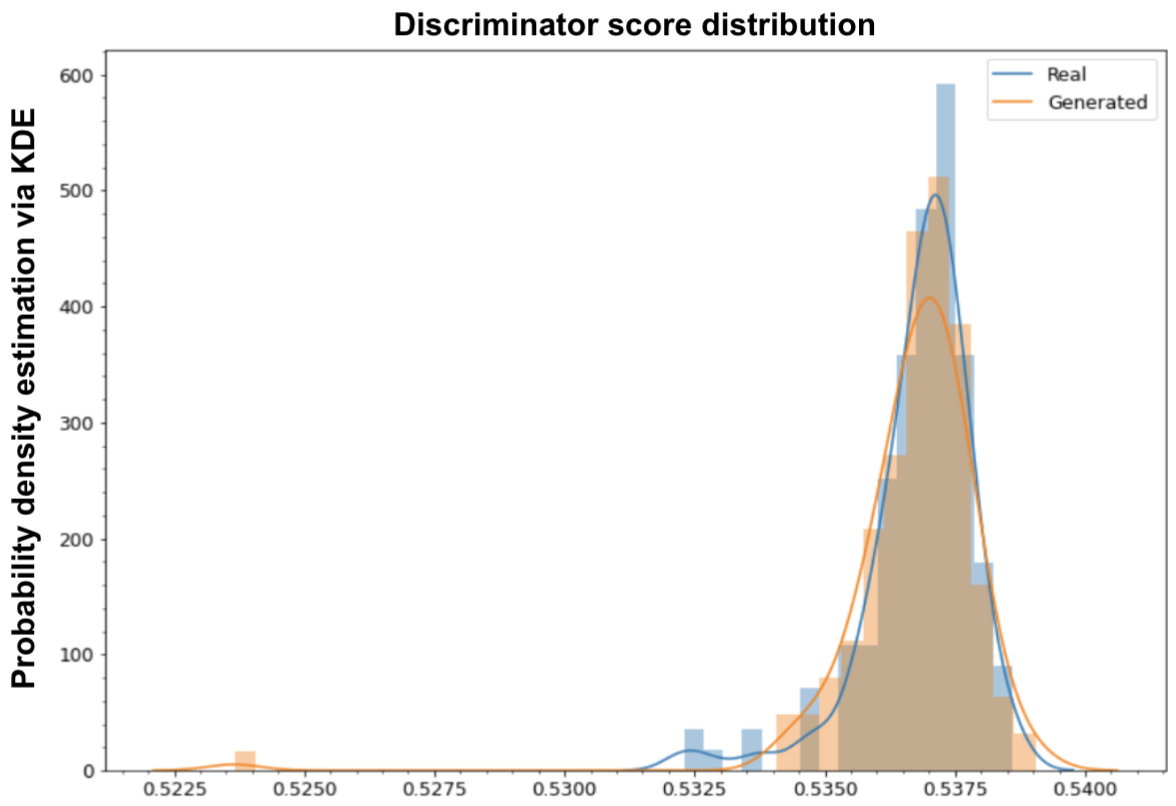}\label{fig:discr_1}}
\caption{Training progression of distributions of discriminator scores (estimates of the probability density functions from the Kernel Density Estimation). At convergence, we observe an overlap between 'real' and 'generated' data discriminator scores distributions, centered around 0.5 with low discrimination variance. We also passed random noise to the discriminator to validate it - 'random' label. }
\label{fig:discr_scores}
\end{figure*}
\begin{figure*}[t!]
\centering
\hspace*{-1cm}  
\subfloat[Historical.]
{\includegraphics[width=0.38\textwidth]{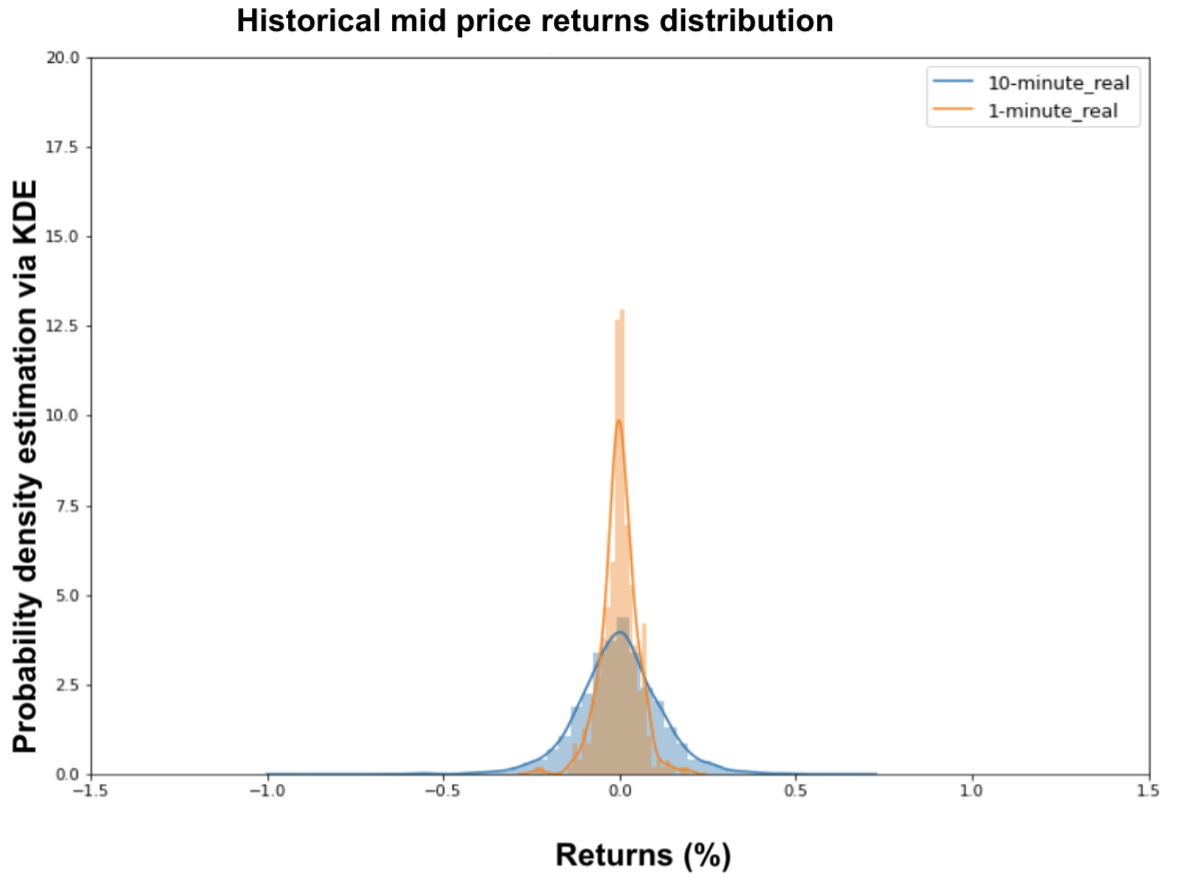}
}
%\hspace*{-2cm}  
\subfloat[GAN-generated after 1000 iterations.]
{\includegraphics[width=0.38\textwidth]{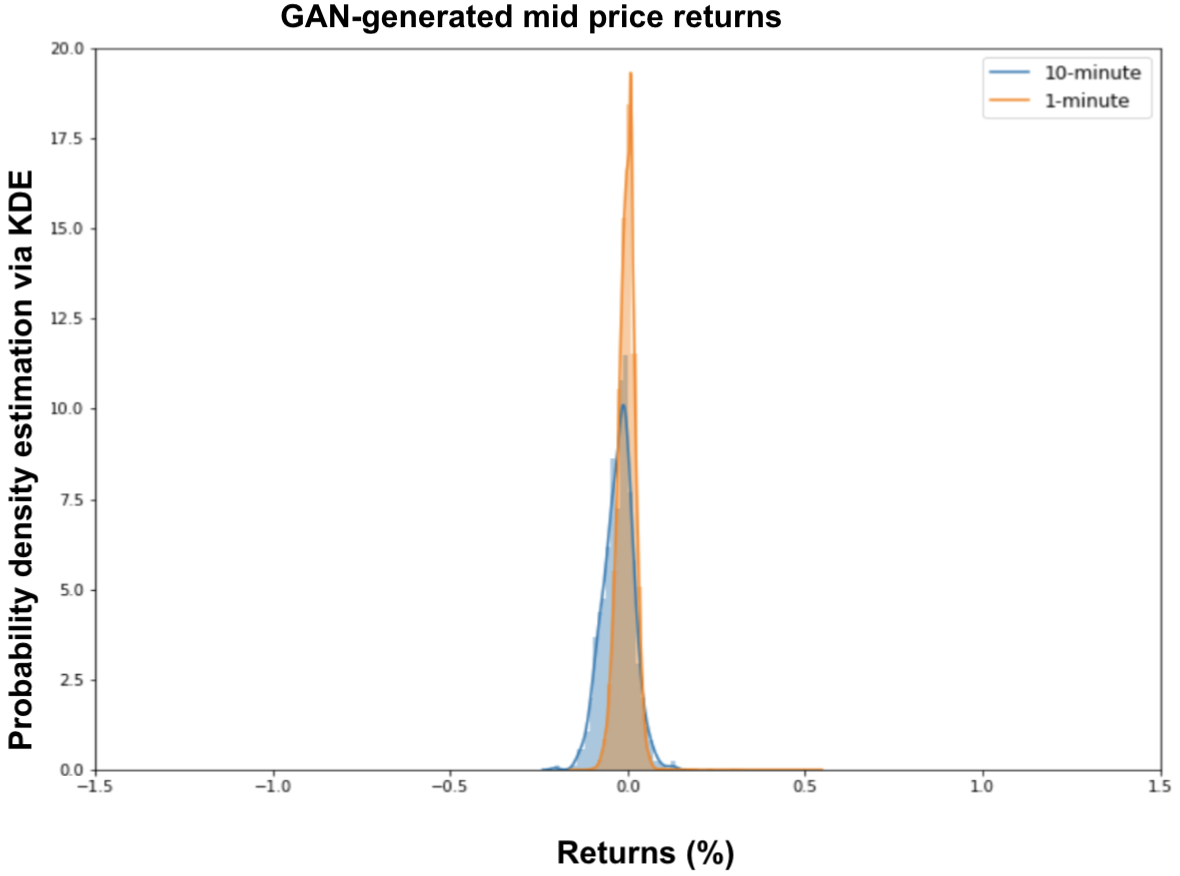}}
%\hspace*{-2cm} 
\subfloat[GAN-generated after 10000 iterations. Note that distributions are similar to (a).]
{\includegraphics[width=0.38\textwidth]{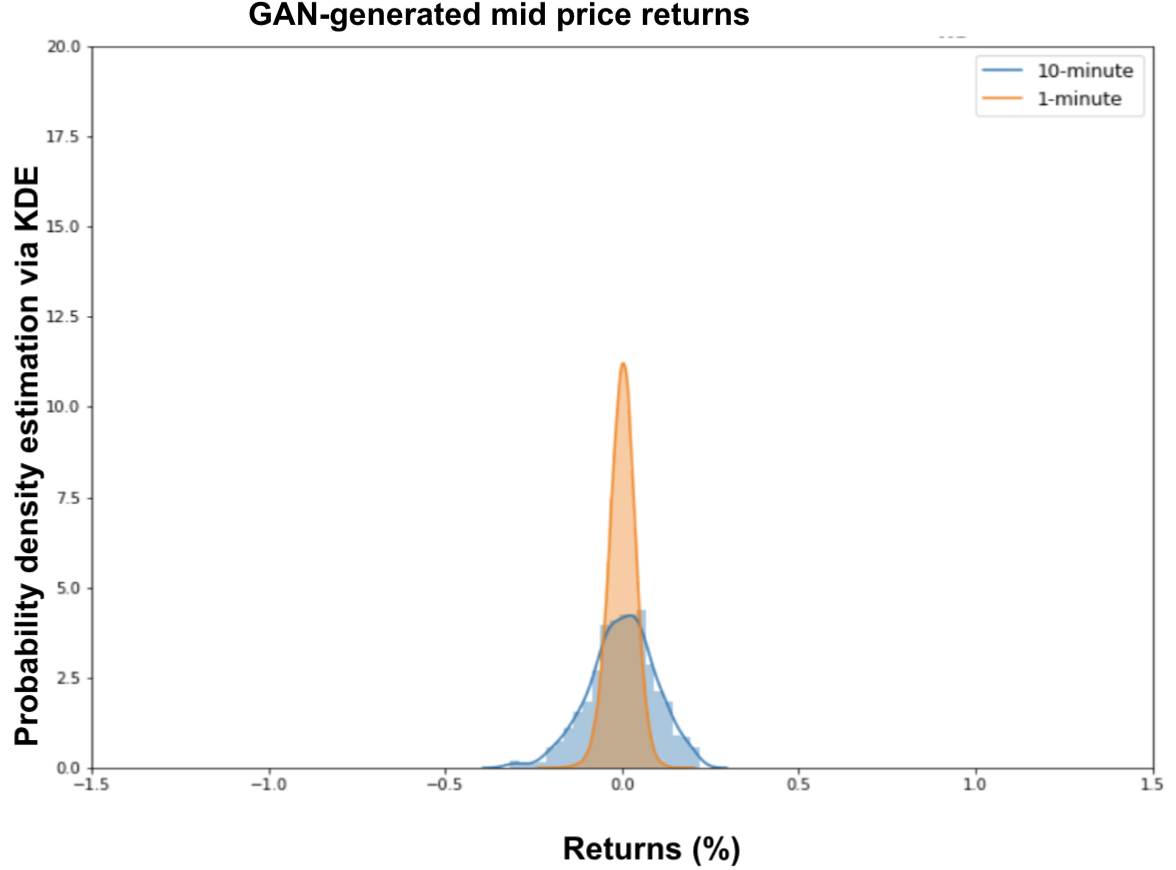}}
\caption{(a) Historical distributions of mid price returns. (b,c) Training progression of 1- and 10-minute mid price return distributions.}
\label{fig:returns}
\end{figure*}

\subsection{1D self-attention layer}
\label{self_attention_appendix}
\begin{figure*}[t!]
\centering
\includegraphics[width=0.4\textwidth]{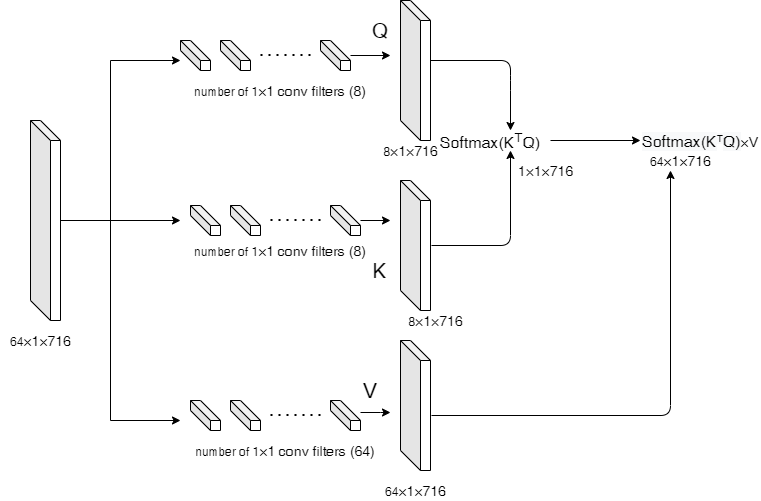}
\caption{The 1D self-attention mechanism first starts with 1 by 1 convolutional layers along the channel dimension for memory efficiency. Then it follows the standard key, query, values scheme applied to the feature map. Kernel size for the feature spaces $f$, $g$ is $8$ and $8$ for feature space $h$.}
\label{fig:self_attention}
\end{figure*}

In Figure~\ref{fig:self_attention} we present in more detail the attention mechanism that is accountable for encoding the global correlations of volumes and mid price returns feature maps. 

\subsection{GAN architecture}
\label{appendix_GAN}

\begin{figure*}[t!]
\centering
\includegraphics[scale=0.4]{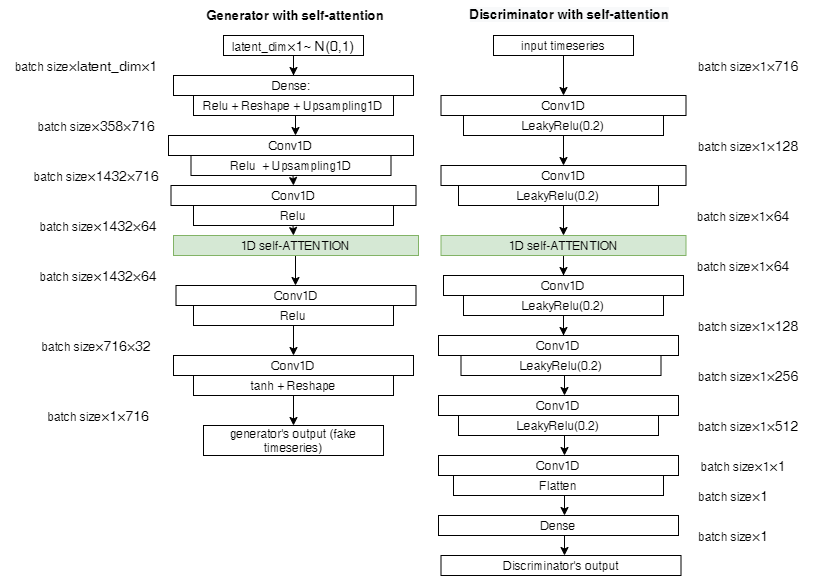}
\caption{Generative adversarial network architecture with self-attention.}
\label{fig:gan_discr}
\end{figure*}

\begin{figure*}[t!]
\centering
%\vspace*{-20cm}
\includegraphics[scale=0.4]{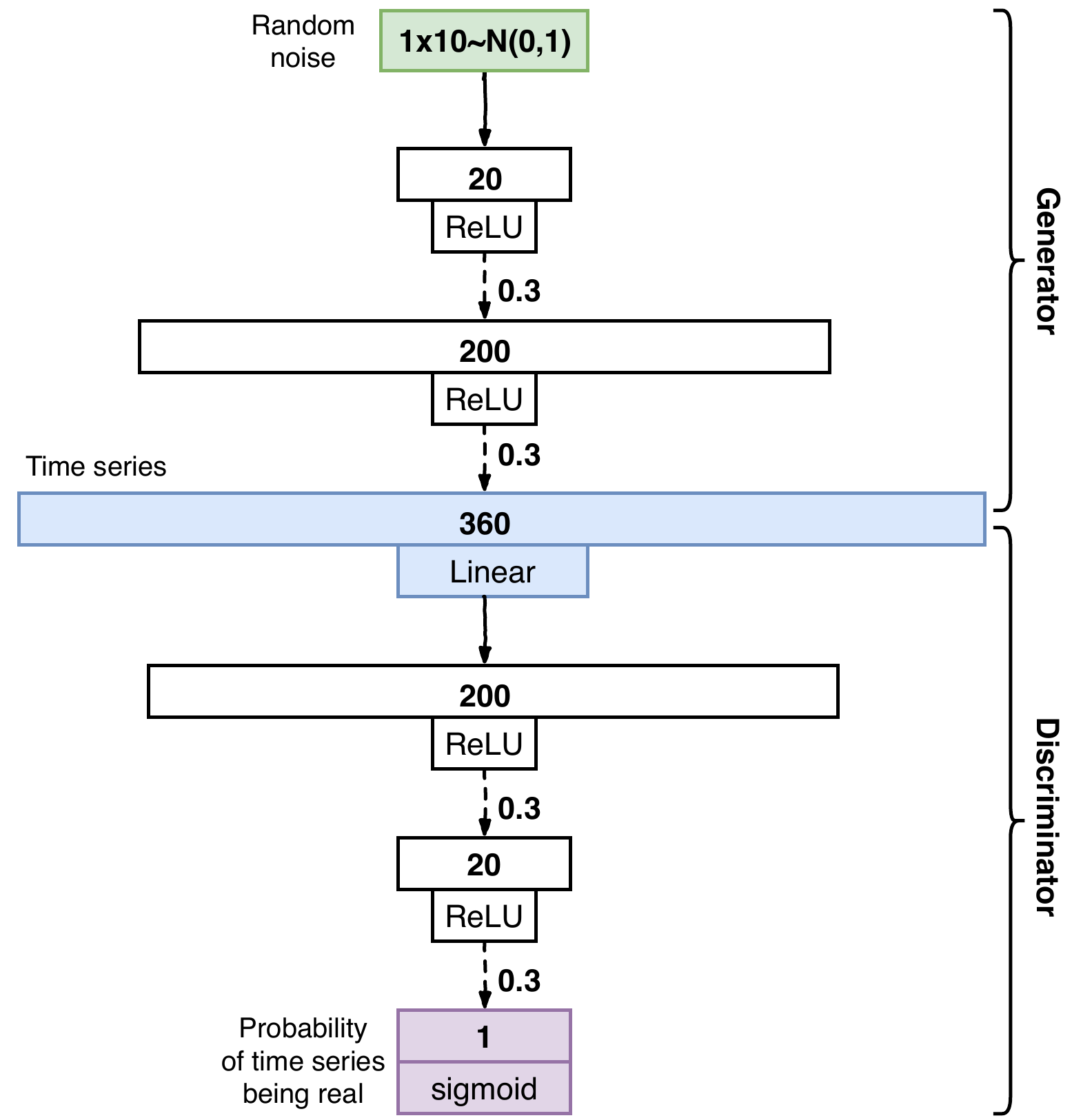}
\caption{Generative adversarial network architecture without self-attention (used for the ablation study). The punctured lines represent dropout layers.}
\label{fig:gan_discr_noattn}
\end{figure*}
\subsection{Ablation study and evaluation}
\begin{figure*}[t!]
%\hspace*{-3.2cm}
\centering
\subfloat[Ablation study: GAN without self-attention. Trained with latent dimension 10 using mid-price returns time series only.]
{\includegraphics[width=0.4\textwidth]{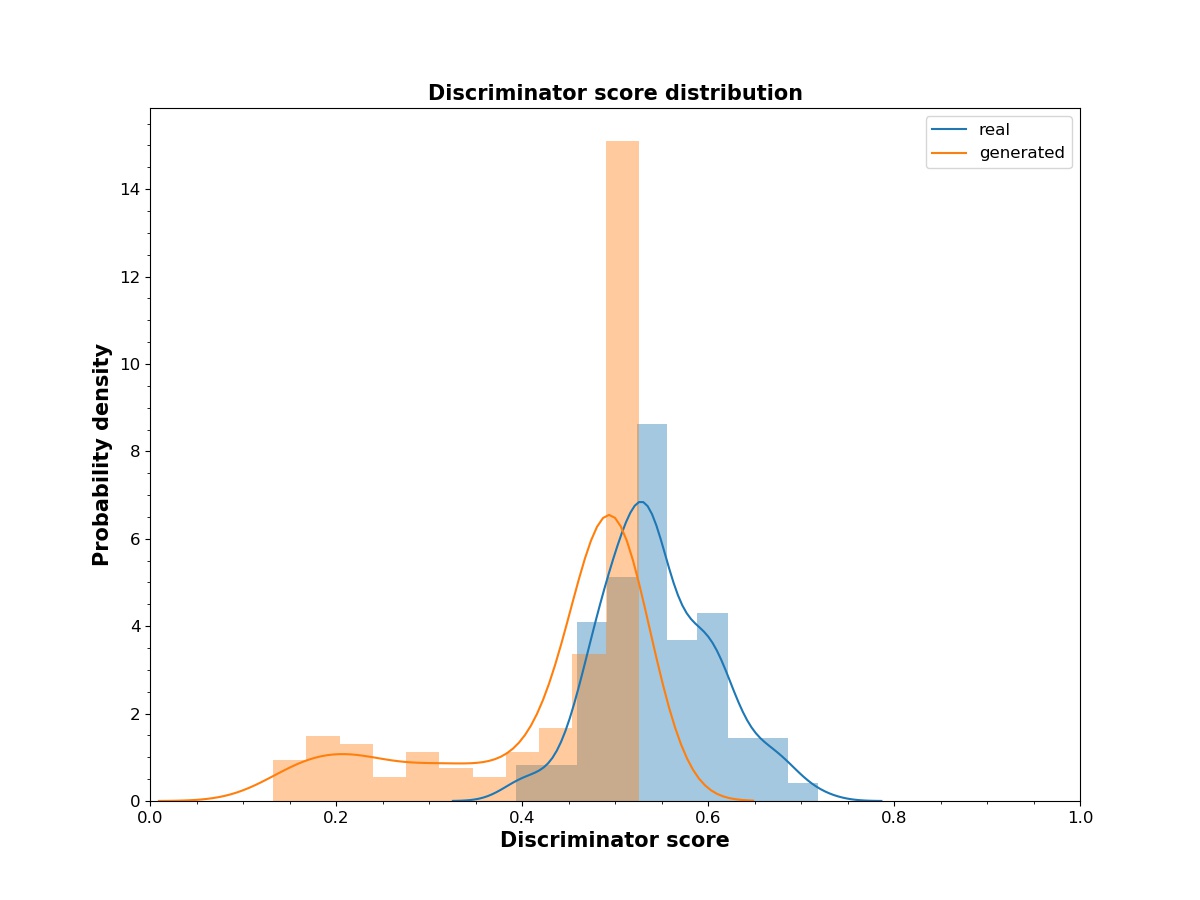}
\label{ablation1}
}
\hspace*{0.1cm}
\subfloat[Ablation study: GAN with self-attention. Trained with latent dimension 10 using mid-price returns time series only.]
{\includegraphics[width=0.4\textwidth]{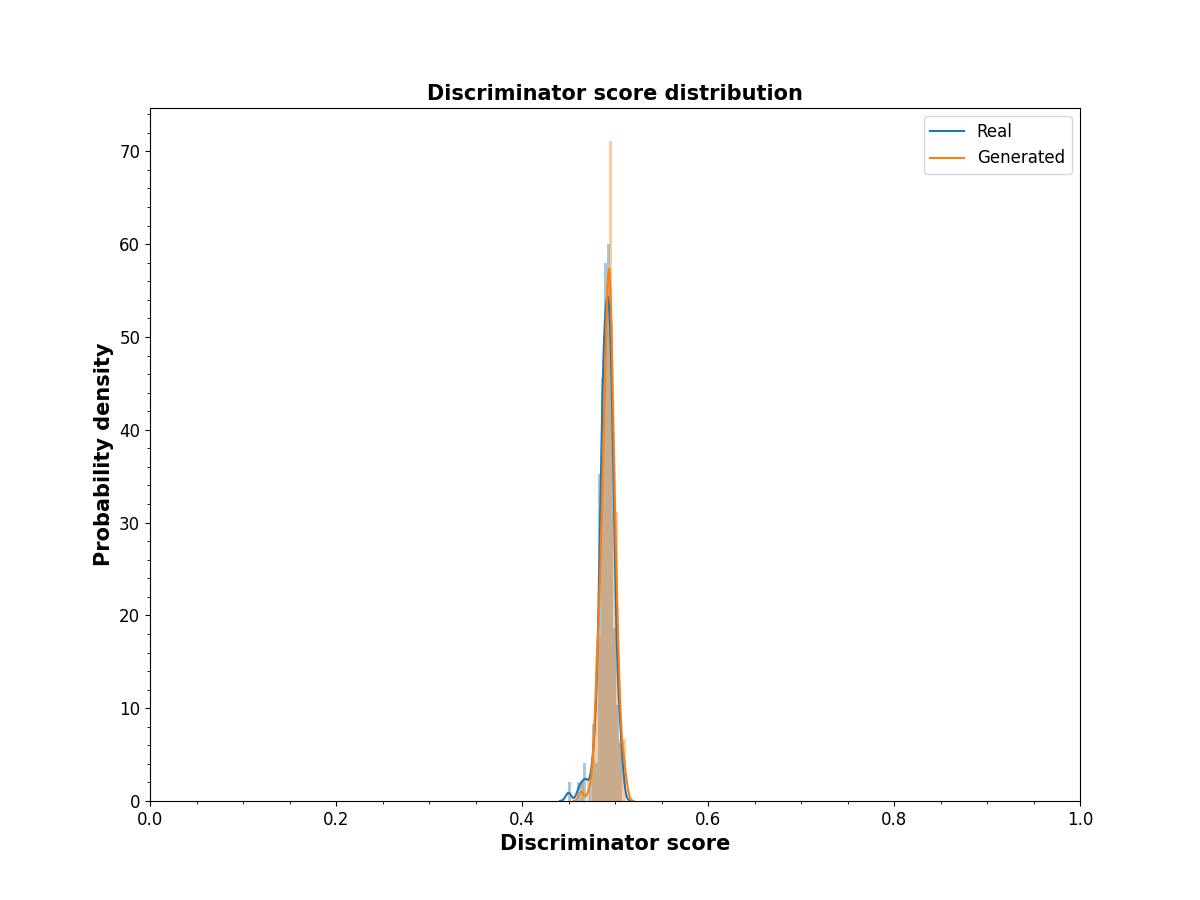}
\label{ablation2}
}
\caption{Adding self-attention to GANs reduces the variance of discriminator score distribution and makes it more concentrated around $\frac{1}{2}$ both for real and generator-produced time series}
\end{figure*}

To justify the complexity of our WGAN with gradient penalty and self-attention architecture in Figure~\ref{fig:gan_discr}, we compare it to a simpler GAN without self-attention that takes same latent dimension as an input, but whose both generator and discriminator are given by feed-forward neural networks with dropout layers and ReLU activation in intermediate layers. This discriminator architecture was studied in \cite{TimeSeriesClass} and is depicted in Figure~\ref{fig:gan_discr_noattn}. We train both GANs on mid price return time series and observe that GANs with self-attention produce discriminators with better recognition capabilities, as evident from Figures~\ref{ablation1} and ~\ref{ablation2}. One can see that adding self-attention to GANs reduces the variance of discriminator score distribution and makes it more concentrated around $\frac{1}{2}$ both for real and generator-produced time series. We further observe that by using both mid price return and volume time series for training a GAN with self-attention, we achieve sufficient discrimination accuracy quicker than by using only mid price returns.

\begin{figure*}[t!]
\centering
\hspace*{0.1cm}
{\includegraphics[width=0.45\textwidth]{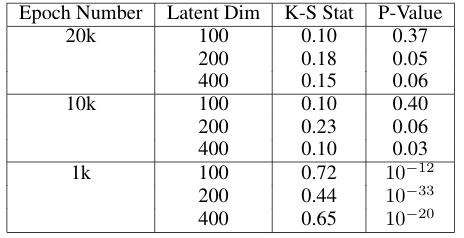}
\label{KStable}
}
\caption{Two sample Kolmogorov-Smirnov test. At 10k iterations for a latent dimension of 100, we cannot distinguish between historical discriminator scores distribution and generated one based on the two sample test.} 
\end{figure*}

To measure the performance of our model  more quantitatively and to study the impact of the latent dimension size, we also conducted analysis using two-sample Kolmogorov-Smirnov test. The historical and generated discriminator scores distributions are compared. The null hypothesis (H0) is that "the two samples are drawn from the same underlying distribution".
From Figure~\ref{KStable}, p-values indicates that after 10k iterations, for latent dimensions equal to 100 and 200 we can't reject the null hypothesis. This means that our generator is producing a distribution of time series that follows the same underlying distribution that the historical. For latent dimension equal to 400, the system requires at least 20k iterations to generalize well and to produce realistic distribution. 

\end{document}